\pdfoutput=1
\documentclass{article}
\usepackage{ijcai19}

\usepackage{times}
\usepackage{soul}
\usepackage{url}
\usepackage{color}
\usepackage[utf8]{inputenc}
\usepackage[small]{caption}
\usepackage{graphicx}
\usepackage{amsmath}
\usepackage{booktabs}
\usepackage{algorithm}
\usepackage{algorithmic}
\usepackage{subfigure}
\usepackage{xcolor}
\usepackage{colortbl,booktabs}
\usepackage{amsfonts}
\usepackage[keeplastbox]{flushend}
\usepackage{bm}
\usepackage[linkcolor=blue, anchorcolor=green]{hyperref}

\begin{document}

\title{One-Shot Texture Retrieval with Global Context Metric}

\author{
	Kai Zhu, Wei Zhai, Zheng-Jun Zha\footnotemark[1], Yang Cao\footnotemark[1]
	\affiliations
	University of Science and Technology of China \emails
	\{zkzy, wzhai056\}@mail.ustc.edu.cn, \{zhazj, forrest\}@ustc.edu.cn
}

\maketitle

\renewcommand{\thefootnote}{\fnsymbol{footnote}}
\footnotetext[1]{Corresponding author}

\begin{abstract}
In this paper, we tackle one-shot texture retrieval: given an example of a new reference texture, detect and segment all the pixels of the same texture category within an arbitrary image. To address this problem, we present an OS-TR network to encode both reference and query image, leading to achieve texture segmentation towards the reference category. Unlike the existing texture encoding methods that integrate CNN with orderless pooling, we propose a directionality-aware module to capture the texture variations at each direction, resulting in spatially invariant representation. To segment new categories given only few examples, we incorporate a self-gating mechanism into relation network to exploit global context information for adjusting per-channel modulation weights of local relation features. Extensive experiments on benchmark texture datasets and real scenarios demonstrate the above-par segmentation performance and robust generalization across domains of our proposed method.
\end{abstract}

\section{Introduction}

As texture refers to the fundamental microstructures of natural image, humans have a strong visual perception of texture, which can not only obtains descriptions of new texture from a small number of training samples (few-shot learning) \cite{sung2018learning}, but also marks such texture regions in the other cluttered scenes (texture segmentation) \cite{cimpoi2015deep}. This suggests that texture features provide a powerful visual prior for comprehensive scene understanding \cite{krishna2017visual}. 

\begin{figure}[t]  
	\centering  
	\includegraphics[width=86mm]{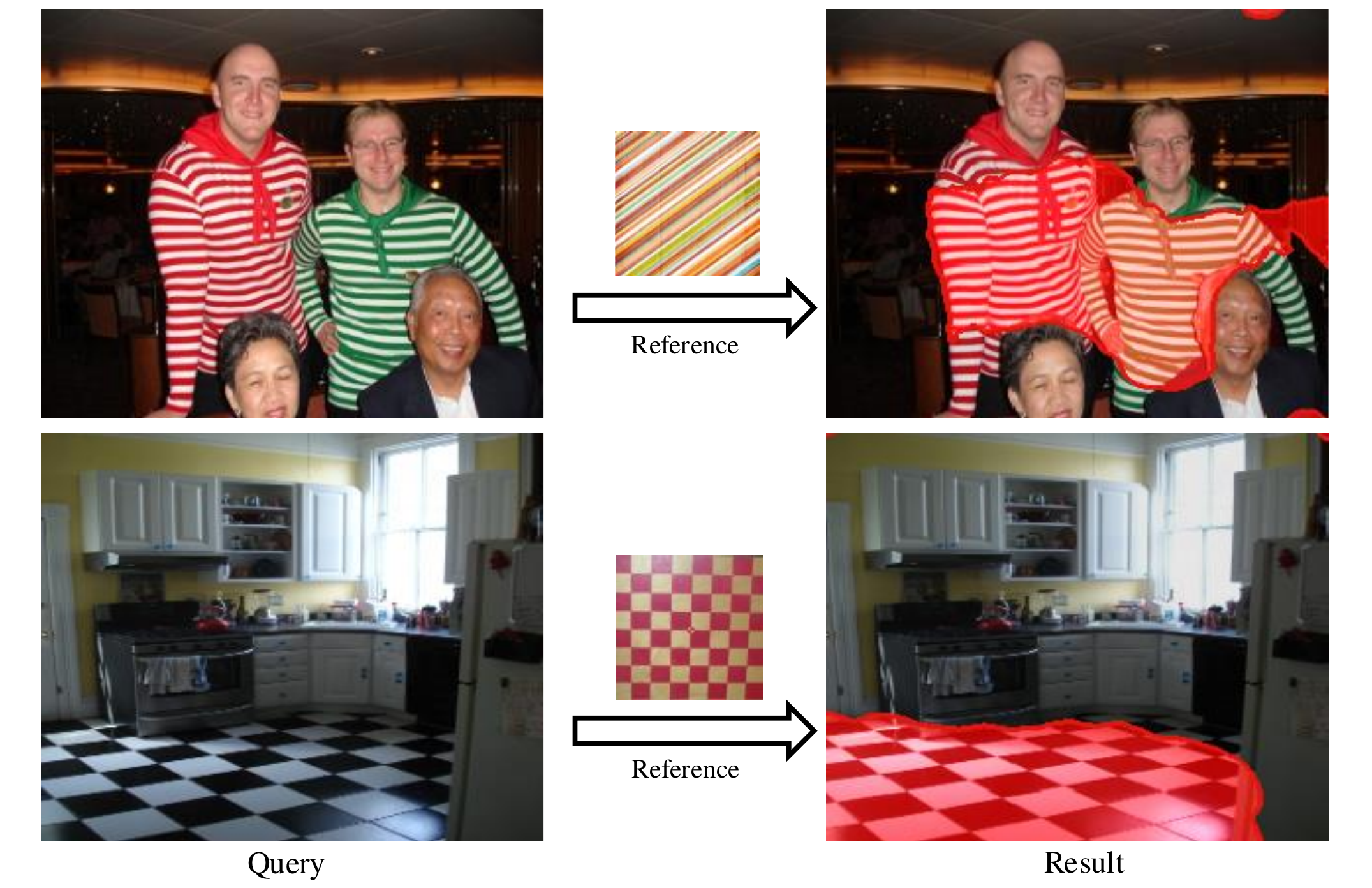}\\  
	\caption{Examples of our task. Given a reference texture of new category, we can segment all pixels of the same texture category within a query image.}  
	\label{example}
\end{figure}

To learn such texture prior, we present the problem of one-shot texture retrieval: given an example of a new reference texture, detect and segment all the pixels of the same texture category within an arbitrary image (see Figure \ref{example}). This task is different from one-shot segmentation of general objects \cite{Shaban2017One}, as the learned texture representation should be invariant to spatial layout but preserve the rough semantic concepts. Therefore, an adaptable and robust texture encoding model should be presented to finely discriminate the orderless texture details. In addition, for texture segmentation, global context is also an important cue since the scene surfaces are usually not completely orderless. A similarity metric should be introduced to balance local spatial details and global scene context for pixel-wise segmentation. 

In this paper, we present an One-Shot Texture Retrieval (OS-TR) network to learn the texture representation, and model the similarities between reference and query image, leading to achieve one-shot texture segmentation towards the reference category. Specifically, a Siamese network \cite{koch2015siamese}is first presented to embed the reference and query image into an encoded representation space by feature learning and parametric prior encoding. Unlike the existing texture encoding methods \cite{xue2018deep} that integrate CNN with orderless pooling, a directionality-aware module is proposed to perceive the local texture variations at each direction, resulting in spatially invariant representation. Then, we propose to incorporate global context into the relation network \cite{sung2018learning} by aggregating feature maps across their spatial dimensions. Different from previous approaches that only consider the similarity of local features or semantic concepts, our method exploits global context information to adjust local relation features by per-channel modulation weights. The key idea is to use a self-gating mechanism for generating global distribution of local relation features with an aggregation unit. The evaluation on synthetic dataset demonstrates the superiority of our proposed model against the state-of-the-art methods \cite{Shaban2017One,Ustyuzhaninov2018One}. Furthermore, the result on nature scenes is promising. 

Our main contribution are as follows:

1.	We introduce a novel one-shot texture segmentation network with global context metric, achieving texture detection and segmentation with a single example of reference texture. 

2.	We propose a directionality-aware module to perceive the local texture variations at each direction, resulting in spatially invariant representation.

3.	We present a global context metric for performing one-shot texture segmentation, which extends relation network with a self-gating mechanism to adjust local relation features.

\section{Related Work}

Our work focuses on one-shot learning, texture modeling and one-shot segmentation task, so in this section we mainly review the research status of these areas. 

\textbf{One-shot learning:} In the computer vision community, one-shot learning has recently received a lot of attention and substantial progress has been made based on metric learning using Siamese neural networks \cite{koch2015siamese,snell2017prototypical}. In addition, there are some work that build upon meta learning \cite{finn2017model}, information retrieval techniques \cite{zhang2012attribute,triantafillou2017few} and generation models \cite{lake2015human} to achieve one-shot learning.

\textbf{Texture representation:} Texture representation is an important research area in computer vision for potential applications in classification, segmentation and synthesis. The research of texture representation are mainly divided two classes: traditional method \cite{kumar2011describable} and CNN-based method \cite{cimpoi2015deep}. Different from object recognition where spatial order is critical for feature representation, texture recognition usually uses an orderless component to provide invariance to spatial layout. 

\textbf{One-shot segmentation:} While the work on one-shot learning is quite extensive, the research on one-shot segmentation \cite{wu2018annotation} have been established only recently, including one-shot image segmentation \cite{Shaban2017One} and one-shot video segmentation \cite{caelles20182018}. The most closely related to our work is \cite{Ustyuzhaninov2018One}, whose task is to segment an input image containing multiple textures by given a patch of a reference texture. Different from their setup in that the reference patch is interactively selected from the input image, our work targets on a more complex problem of one-shot texture retrieval: given an example of a new reference texture, detect and segment all the pixels of the same texture category within an arbitrary image.

\begin{figure*}[t]  
	\centering  
	\includegraphics[width=150mm]{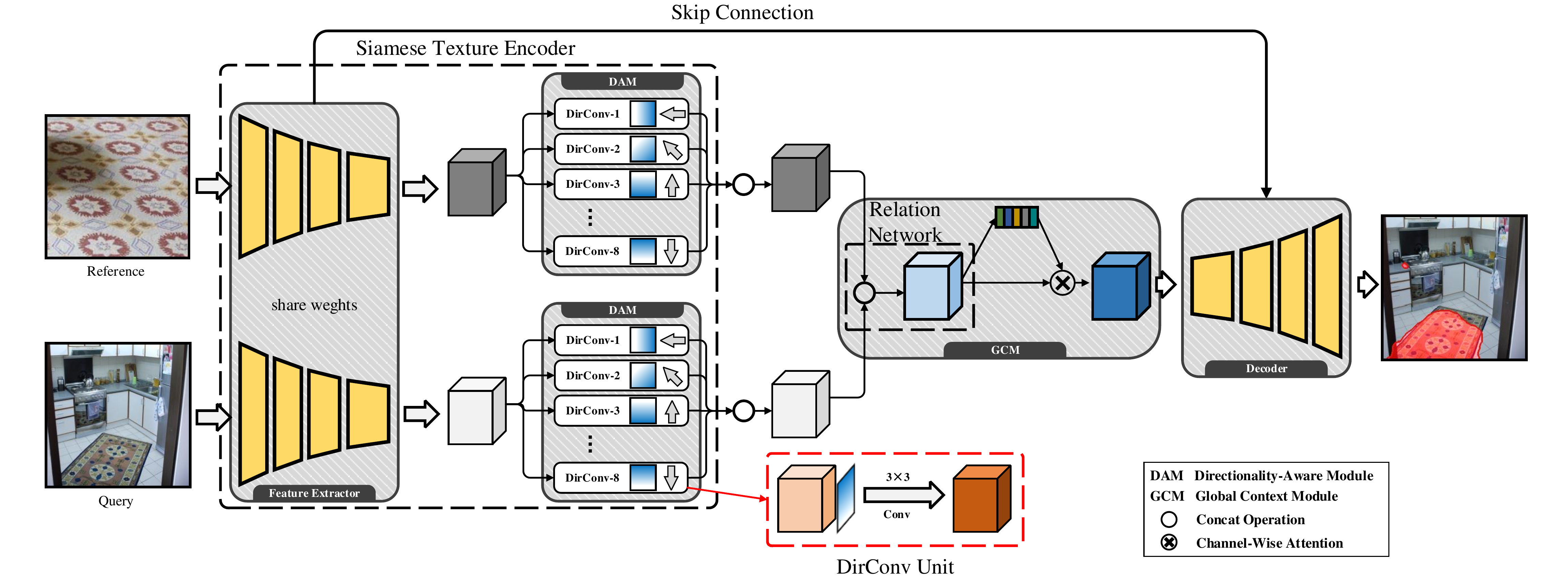}\\  
	\caption{Overview of the OS-TR network. The texture features of the query image and reference are first extracted by the proposal Siamese texture encoder. Through a global context metric, the relation score is obtained by means of global information and then combined with the encoding features to get the final segmentation results.}  
	\label{fig:env1}  
\end{figure*}

\section{One-shot Texture Segmentation}

\subsection{Probelm Setup}

We define a triple $Tr_{i}^{j} = (Q_{i}, R_{i}^{j}, T_{i}^{j})$ and a relation function $F:A_{i}^{j}=F(Q_{i}, R_{i}^{j};\theta)$, where $Q_{i}$ is the $i^{th}$ query image with multiple classes of textures, $T_{i}^{j}$ is the pixel-wise labels corresponding to the $j^{th}$ class texture in $Q_{i}$, $R_{i}^{j}$ is a reference texture image of the same class j, $A_{i}^{j}$ is the actual segmentation result, and $\theta$ is all parameters to be optimized in function F. Our task is to randomly sample triples from the dataset, train and optimize $\theta$, thus minimizing the loss function $L$:
\begin{equation}
\theta _{\ast } = arg \min\limits_{\theta} L(A_{i}^{j}, T_{i}^{j}).
\end{equation}
We expect that the relationship function $F$ can segment the same class texture region in another target image each time it sees a reference texture picture of new class. This is the embodiment of the meaning of one-shot segmentation. It should be mentioned that the texture classes sampled by the test set are not present in the training set, that is, $U_{train} \bigcap U_{test}=\O $. The relation function $F$ in this problem is implemented by the model detailed in Section \ref{subsection1}.

\subsection{Model Architecture}
\label{subsection1}

In this section we will detail the overall framework of the proposed OS-TR network. As shown in Figure \ref{fig:env1}, the network is based on a classic encoder-decoder network \cite{ronneberger2015u} to complete the segmentation task. It consists of the texture encoder for captioning the characteristics of the texture and global context metric that better determines the matching pixels. These two modules will be explained further in Section \ref{texture} and \ref{subsection3}.

The network uses the texture encoders $f_{t}$ to transform the input of two branches into the variable embeddings \cite{zhang2014robust} respectively. The first branch takes the reference texture $R_{i}^{j}$ from the triple as input. And the second branch takes the synthetic texture image as input during training, which may be from a wide range of sources in the real scenarios. We use the texture encoder to perceive the local texture features at each dirction, and the corresponding feature map pairs $M_{1}$, $M_{2}$ obtained can be expressed as follows:
\begin{align}
M_{1}=f_{t}(R_{i}^{j};\theta_{t}),\\
M_{2}=f_{t}(Q_{i}^{j};\theta_{t}).
\end{align}
Here $\theta_{t}$ is the learnable parameter of our texture encoder.

Different from the existing work, one-shot relation network \cite{sung2018learning} proposes a deep nonlinear metric. However, this nonlinaer comparator is limited to local ralation features. Instead we learn a global context metric $g_{c}$ to compare the pixel information of query images with the reference texture by taking global context in consideration. It provides important features for subsequent pixel-level segmentation. In our network,
\begin{equation}
\label{S}
S=g_{c}(M_{1},M_{2};\theta_{c}),
\end{equation}
where S refers to the element-wise relation score, and $\theta_{c}$ is the parameter to be optimized in $g_{c}$.

To generate the same size as the original image, we combine the score $S$ with the information of encoder $f_{t}$ in multiple stages. We make full use of extracted features at different scales in the encoding process to form a refined decoding layer $g_{d}$. The final actual segmentation result $A$ of the original image is obtained through a sigmoid function:
\begin{equation}
A=Sigmoid(g_{d}(S,f_{t_{2}};\theta_{d})).
\end{equation}
Similarly, $\theta_{d}$ stands for the parameters of decoding part.

The number of positive and negative samples in the training set is unbalanced (i.e., the foreground and background of query images). To ensure their equal contribution to optimization function, we use weighted Binary Cross Entory Loss function, that is
\begin{equation}
\begin{split}
L = -\alpha \sum_{p\in T_{+}} logPr(y_{p}=1\mid Q_{i}, R_{i}^{j}; \theta)-\\
(1-\alpha )\sum_{p\in T_{-}} logPr(y_{p}=0\mid Q_{i}, R_{i}^{j}; \theta),
\end{split}
\end{equation}
where $y_{p}$ stands for the ground truth of corresponding pixel p, $\alpha =\left | T_{-} \right |/(\left | T_{-} \right |+\left | T_{+} \right |)$. $T_{+}$ and $T_{-}$ denotes positive and negtive sample sets in training images, respectively. We can see $\theta$ in the loss function is the collection of $\theta_{t},\theta_{c}$ and $\theta_{d}$ described earlier. The purpose of reducing loss function is to optimize the parameters of corresponding modules.


\begin{table}[t]
	\centering
	\small
	\begin{tabular}{|c|c|c|c|}
		\hline
		Layer & Input & Output & Type \\ \hline
		\multicolumn{4}{|c|}{\cellcolor[HTML]{C0C0C0}Feature Extractor} \\ \hline
		\multicolumn{4}{|c|}{ResNet-50: $conv1-conv5$} \\ \hline
		\multicolumn{4}{|c|}{\cellcolor[HTML]{C0C0C0}Directionality-Aware Module} \\ \hline
		\begin{tabular}[c]{@{}c@{}}DirConv\\ unit\end{tabular} & \begin{tabular}[c]{@{}c@{}}$8\times 8\times$ \\$(2048+\textbf{1})$\end{tabular} & \begin{tabular}[c]{@{}c@{}}$8\times 8$\\ $\times 256$\end{tabular} & $3\times 3$ Conv \\ \hline
		Join & \begin{tabular}[c]{@{}c@{}}8$\times$\\ {[}$8\times 8\times 256${]}\end{tabular} & \begin{tabular}[c]{@{}c@{}}$8\times 8$ \\ $\times 2048$\end{tabular} & Cat \\ \hline
		\multicolumn{4}{|c|}{\cellcolor[HTML]{C0C0C0}Global Context Metric} \\ \hline
		Join & \begin{tabular}[c]{@{}c@{}}2$\times$\\ {[}$8\times 8\times 2048${]}\end{tabular} & \begin{tabular}[c]{@{}c@{}}$8\times 8$\\ $\times 1024$\end{tabular}  & \begin{tabular}[c]{@{}c@{}}Cat+\\ $3\times 3$ Conv\end{tabular} \\ \hline
		Aggregation & $8\times 8\times 1024$  & \begin{tabular}[c]{@{}c@{}}$8\times 8$\\ $\times 1024$\end{tabular}  & \begin{tabular}[c]{@{}c@{}}Max pooling\\ + $1\times 1$ Conv\end{tabular} \\ \hline
		Weighting & $8\times 8\times 1024$  &  \begin{tabular}[c]{@{}c@{}}$8\times 8$\\ $\times 1024$\end{tabular}  & \begin{tabular}[c]{@{}c@{}}Channel-wise\\ Multiplication\end{tabular} \\ \hline
		\multicolumn{4}{|c|}{\cellcolor[HTML]{C0C0C0}Decoder} \\ \hline
		\begin{tabular}[c]{@{}c@{}}Upsample\\ ($\times 4$)\end{tabular} & $8\times 8\times 1024$  & \begin{tabular}[c]{@{}c@{}}$256\times 256$\\ $\times 1$\end{tabular} & \begin{tabular}[c]{@{}c@{}}Bilinear\\ Interpolation\\ + $1\times 1$ Conv\\ + Sigmoid\end{tabular} \\ \hline
	\end{tabular}
	\caption{Details of our network. `$\times 4$' represents four upsampling operations. The bold $\textbf{1}$ in the DirConv unit represents a directional map.}
	\label{table}
\end{table}

\subsection{Texture Encoder}
\label{texture}

To perceive the local texture variations at each direction, we propose the texture encoder consisting of feature extraction and directionality-aware module. As shown in Figure \ref{fig:env1}, ResNet (ResNet in this paper represents ResNet-50 from \cite{he2016deep} that removes the last fully connected layer) is used to extract preliminary features $P_{1}$, $P_{2}$ from the query image and reference texture. Then the DirConv unit $g_{d}$ is presented to capture eight directional texture features $G_{i}$ under the guidance of corresponding directional feature map $D_{i}$. We only detail the first branch as the parameters and structures of two branches are identical. Specifically,
\begin{equation}
G_{i} = g_{d}(cat (P_{1},D_{i})) (i=1,2\cdots8),
\end{equation}
where $g_{d}$ denotes a convolution block which contains a $3\times 3$ convolution, a batch normalization and a ReLU activation layer, cat refers to the concatenation function in the channel dimension. In this paper, eight directions refer to top, bottom, left, right, top left, bottom left, top right, and bottom right. Each directional map $D_{m}$ is a generated trend graph that decreases from $1$ to $0$ along a certain direction. Finally, the output features of different branches are concatenated to form the whole spatial invariant feature $M_{1}$, that is, 
\begin{equation}
M_{1}= g_{h}(cat (G_{1},G_{2}\cdots G_{8})),
\end{equation}
where $g_{h}$ represents a set of standard convolution block. Since the proposed DirConv unit is sensitive to the local variation of image along each direction, it can provide the network with good adaptability to spatial distortion and scale changes. 

\begin{table}
	\centering
	\begin{tabular}{cc}
		\toprule[1pt]
		\toprule[1pt]
		$U_{test_{i}}$  & classes \\
		\midrule
		i=0       & perforated, pitted, pleated, polka-dotted, porous     \\
		i=1       & stained, stratified, striped, studded, swirly     \\
		i=2       & veined, waffled, woven, wrinkled, zigzagged     \\
		\bottomrule[1pt]
		\bottomrule[1pt]
	\end{tabular}
	\caption{Specific class names of three test sets defined in section 4.1.}
	\label{tab:plain1}
\end{table}

\subsection{Global Context Metric}
\label{subsection3}

To achieve pixel-wise segmentation, we incorporate global context into local relation feature to adjust per-channel modulation weights. Firstly, we use a non-linear function $g_{m}$ to capture local relation features L:
\begin{equation}
L = g_{m}(cat (M_{1},M_{2})), L \in \mathbb{R}^{H\times W\times C}
\end{equation}
where $g_{m}$ represents two sets of standard convolution blocks, $H\times W$ and $C$ refers to spatial and channel dimensions of the feature map, respectively. It can be seen that the two-layer convolution module represents more metric possibilities, which is not limited by cosine, Euclidean distance \cite{Vinyals2016Matching,snell2017prototypical}, etc. However, it only considers local feature similarity which has its limitation. Instead we take global context into consideration by aggregating feature maps across their spatial dimensions, which is similar to SENet \cite{hu2018squeeze}.

Let $L = \left [ l_{1},l_{2}\cdots l_{C} \right ](l_{i}\in \mathbb{R}^{H\times W})$ denote the local relation features of different channels. In this paper, we simply obtain the global information of each channel $\beta \in \mathbb{R}^{C}$ through maximum pooling $g_{max}$. The aggregation unit can be represented as follows: 
\begin{equation}
\beta _{i}=g_{max}(l_{i}), i=1,2\cdots C.
\end{equation}
Next we use the collected global context to balance the relation features. Here we use two simple $1\times 1$ convolutional blocks $g_{s}$ to achieve per-channel modulation weights. Finally, the obtained weights are normalized to 0-1 and used as multiplication coefficient of corresponding channel of original feature $L$. The re-layout of local relation features can be formulated as follows:
\begin{equation}
S=\gamma \cdot L=sigmoid({g_{s}(\beta )})\cdot L,
\end{equation}
where $S$ is the same as the one in Equation \ref{S} and $\cdot$ represents the channel-wise multiplication. It can be seen from the visualization of experimental part that global context metric realizes the fine adjustment of different texture matching pairs. It is beneficial for the optimization of the final segmentation result.

\begin{table}[t]
	\centering
	\setlength{\tabcolsep}{4mm}{
		\begin{tabular}{lccccc}
			\toprule[1pt]
			\toprule[1pt]
			Method  & i=0 & i=1 & i=2 & mean \\
			\midrule
			Baseline  & 56.4  & 47.2 & 44.9 & 49.5   \\
			+Tex   & 59.3  & 50.3 & \textbf{46.4} & 52.0   \\
			+Glo   & 59.4  & 51.5 & 45.4 & 52.1   \\
			+Tex\& Glo  &\textbf{60.7}  & \textbf{52.8} & 44.8 & \textbf{52.8}   \\
			\bottomrule[1pt]
			\bottomrule[1pt]
	\end{tabular}}
	\caption{Results of ablation study. `Tex' and `Glo' represent texture encoder and global context metric respectively. The middle three columns represent the mean IoU metric(\%) for the three test sets, and the mean represents the average of results for the three test sets.}
	\label{tab:plain2}
\end{table}

\begin{figure*}[t]
	\centering  
	\includegraphics[width=150mm]{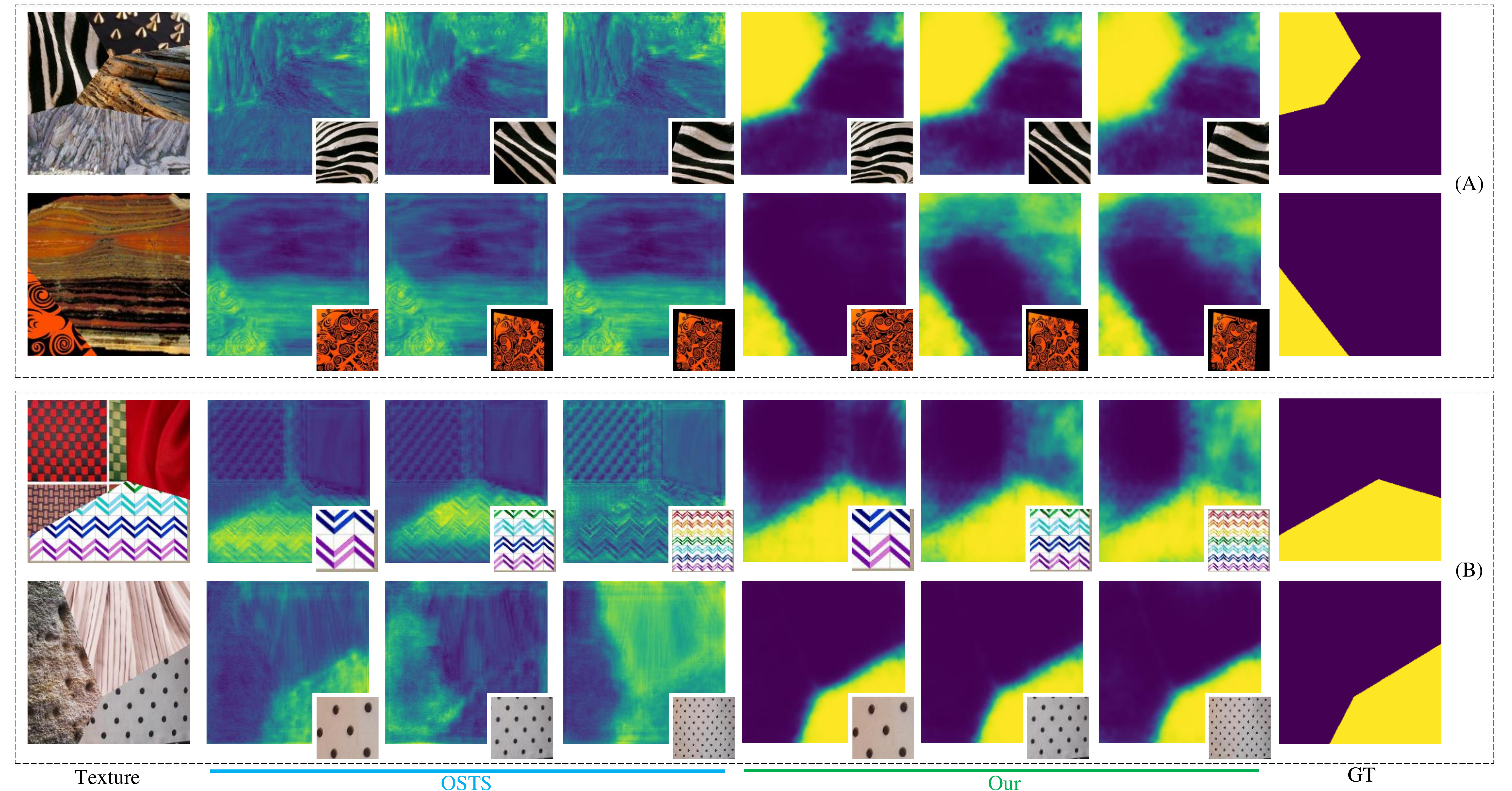}\\  
	\caption{Verification results of spatial invariance. Part (A) and (B) represent the results of affine transformation and scale change respectively. From left to right, they represent query image, results of OSTC compared to ours and groud truth respectively. Reference textures are shown in the lower-right corner of each image of results.}   
	\label{fig:env4}
\end{figure*}

\section{Experiments}

To validate the superiority of our model in one-shot texture segmentation task, we designed a series of experiments based on Describable Textures Dataset(DTD) \cite{Cimpoi2014Describing} dataset. In this section, we first introduce the preprocessing process of the dataset, and then perform ablation experiments on main modules in the model. While demonstrating its superiority from objective indicators, we also give some good visualization results. Finally, we compare our model to the most advanced model One-Shot Learning for Semantic Segmentation (OSLSM) \cite{Shaban2017One} and One-shot Texture Segmentation (OSTC) \cite{Ustyuzhaninov2018One} in the current one-shot segmentation field. We also introduce expanded experimental contents in supplementary materials\footnote{\url{https://github.com/zhukaii/ijcai2019}}.

\subsection{Dataset and Setting}

To solve the proposed one-shot texture segmentation task, we redivide DTD dataset into training and test sets. We divide the last 15 classes of DTD dataset into three test subsets on average, and the remaining classes are used as the training set. The specific class name is shown in Table \ref{tab:plain1}. During the training, we randomly sample 2-5 texture images from the training set to synthesize query images (these textures may come from the same class), and generate labels for one of the texture regions. The technique of texture synthesis comes from \cite{Ustyuzhaninov2018One}. Finally, we randomly sample the reference texture image from the training set with the same kind of labels to form the triple defined in Section 4.1. In the test phase, we synthesize $240$ query images (a total of 960 images) from the three subtest sets in the same way. We send them to the trained model to calculate evaluation indicators and then average them to ensure that different texture classes in the test set are relatively fair. The regular evaluation indicator IoU is used in our task.

Our model uses the SGD optimizer during the training process. The initial learning rate is set to $0.001$ and the attenuation rate is set to $0.0005$. The model stops training after $1000$ epochs, where each epoch synthesizes $240$ query images. All images are resized to $256\times 256$ size and the batch size is set to $16$.

\subsection{Ablation Study} 

We conduct several ablation experiments to prove the effectiveness of directionality-aware module and global context metric of our model in Table \ref{tab:plain2}. We first train a baseline without the two modules, and then add them respectively to compare the results. First of all, we can see that the global context metric improves the model by $2.6\%$ mean IoU. This is due to its use of global information to help the model adjust the overall feature distribution. In order to explain this more vividly, we make a visualization of per-channel modulation weight distribution learned by global context metric in Section \ref{subsection2}. Then, the result has a $2.5\%$ mean IoU boost with the dierctionality-aware module. This module takes into account the characteristics of texture spatial distribution, which is very helpful for texture encoding. The specific analytical experiment is also shown in Section \ref{subsection2}. Finally, we add both the two modules to form our OS-TR network, which has a $3.3\%$ improvement over the baseline.

\subsection{Evaluation}
\label{subsection2}

\textbf{Spatial Invariance:} 
To evaluate the performance of our model on the spatial invariant texture representation, we conduct the following experiments. First, in order to demonstrate the adaptability of our model to spatial distortion, we show its segmentation effect by affining the reference texture image. As shown in Figure \ref{fig:env4} (A), we take several query images as examples and randomly determine the parameters of affine transformation to compare the segmentation results. It can be seen the results achieve the segmentation effect similar to the original reference texture. As shown in Figure \ref{fig:env4} (B), we select three scale reference textures $256\times 256$, $128\times 128$ and $64\times 64$ to evaluate the sensitivity of the model to scale change. The segmentation effect is still excellent. It is the proposed DirConv unit that is sensitive to the local variation of image along each direction, so that the spatial arrangement of texture can be accurately extracted.

\begin{figure}[t]
	\centering
	\subfigure{
		\begin{minipage}[t]{0.32\linewidth}
			\centering
			\includegraphics[width=1.1in]{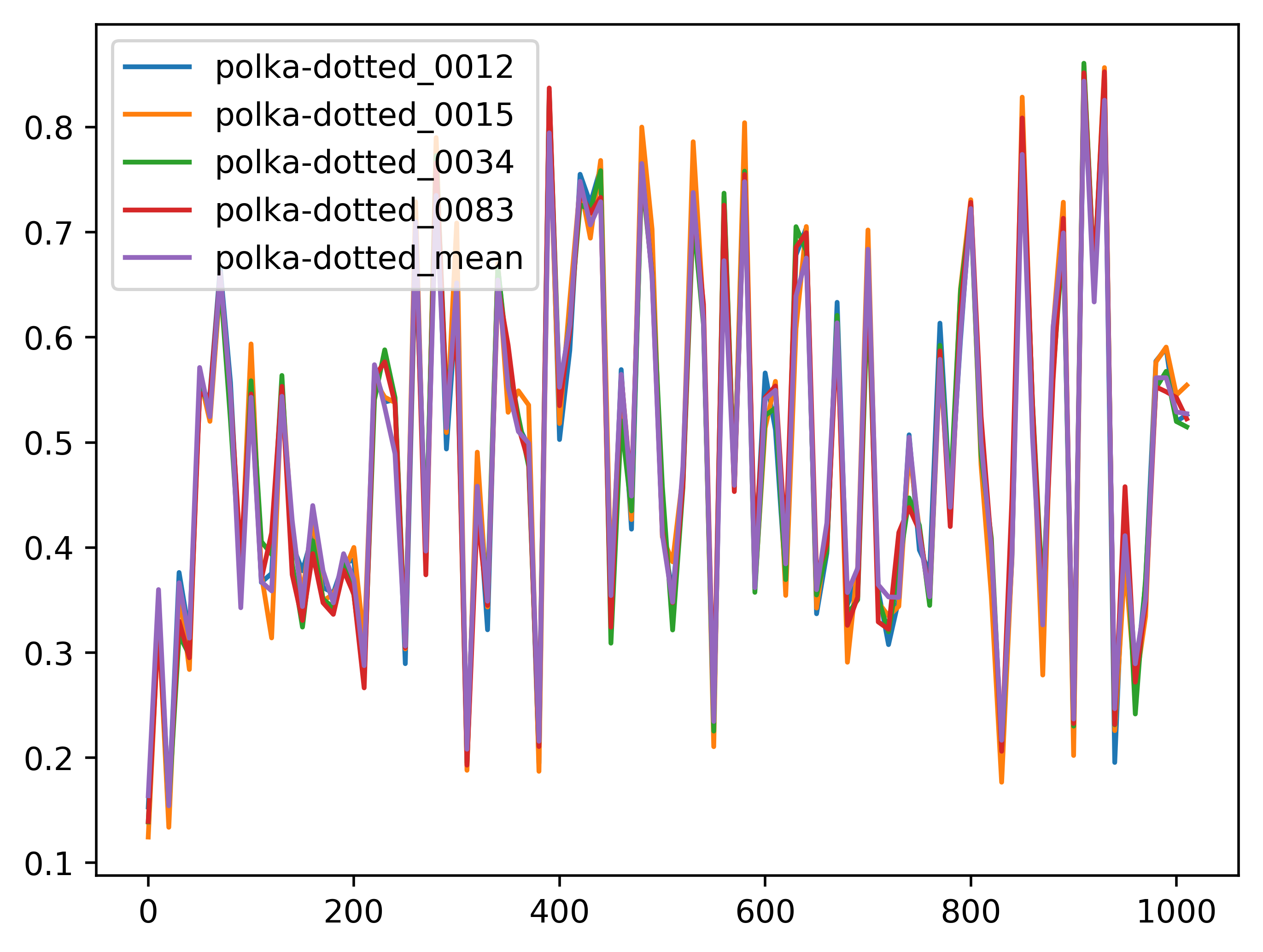}
			%\caption{fig1}
		\end{minipage}%
	}%
	\subfigure{
		\begin{minipage}[t]{0.32\linewidth}
			\centering
			\includegraphics[width=1.11in]{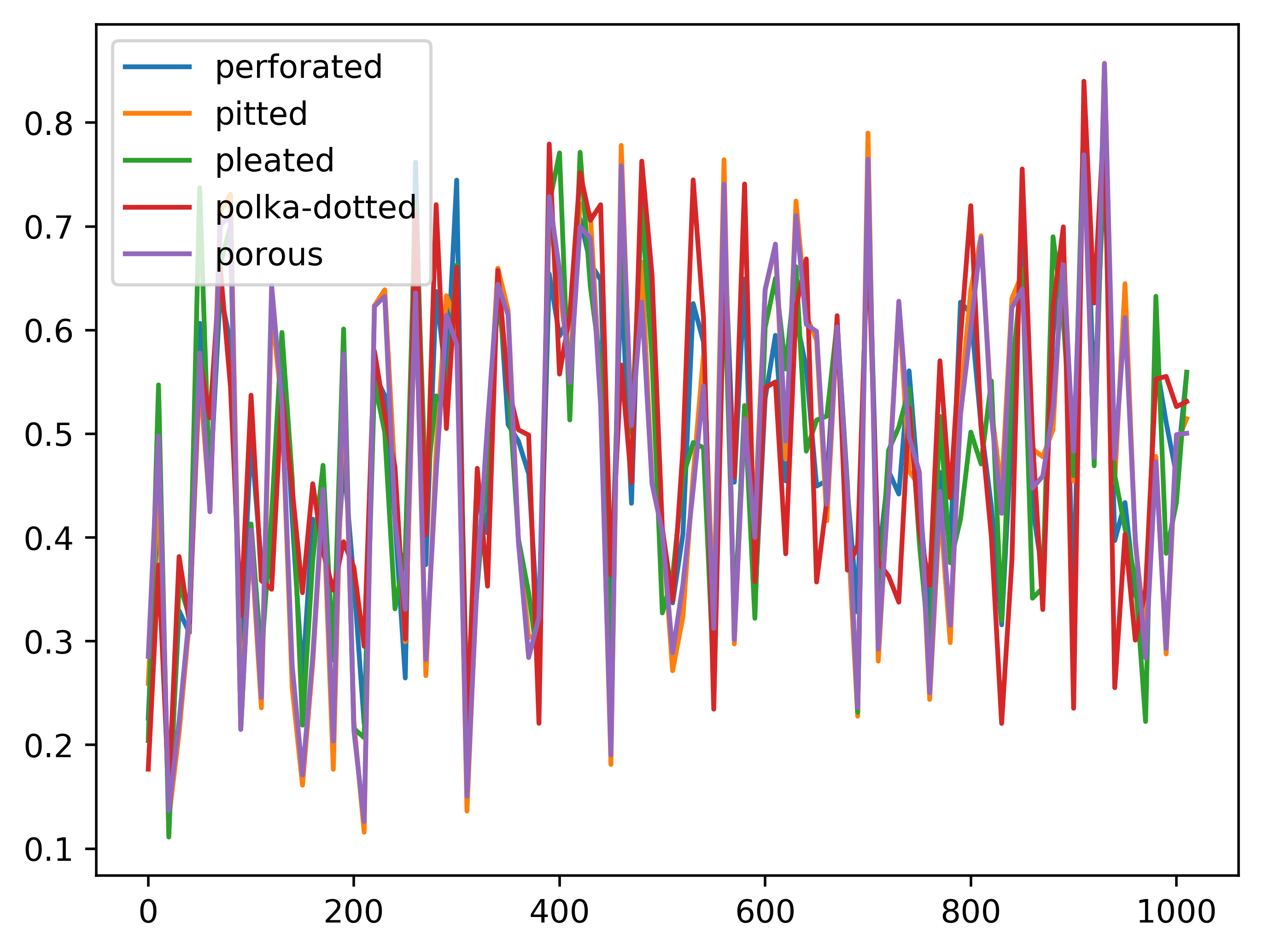}
			%\caption{fig2}
		\end{minipage}%
	}%
	\subfigure{
		\begin{minipage}[t]{0.32\linewidth}
			\centering
			\includegraphics[width=1.1in]{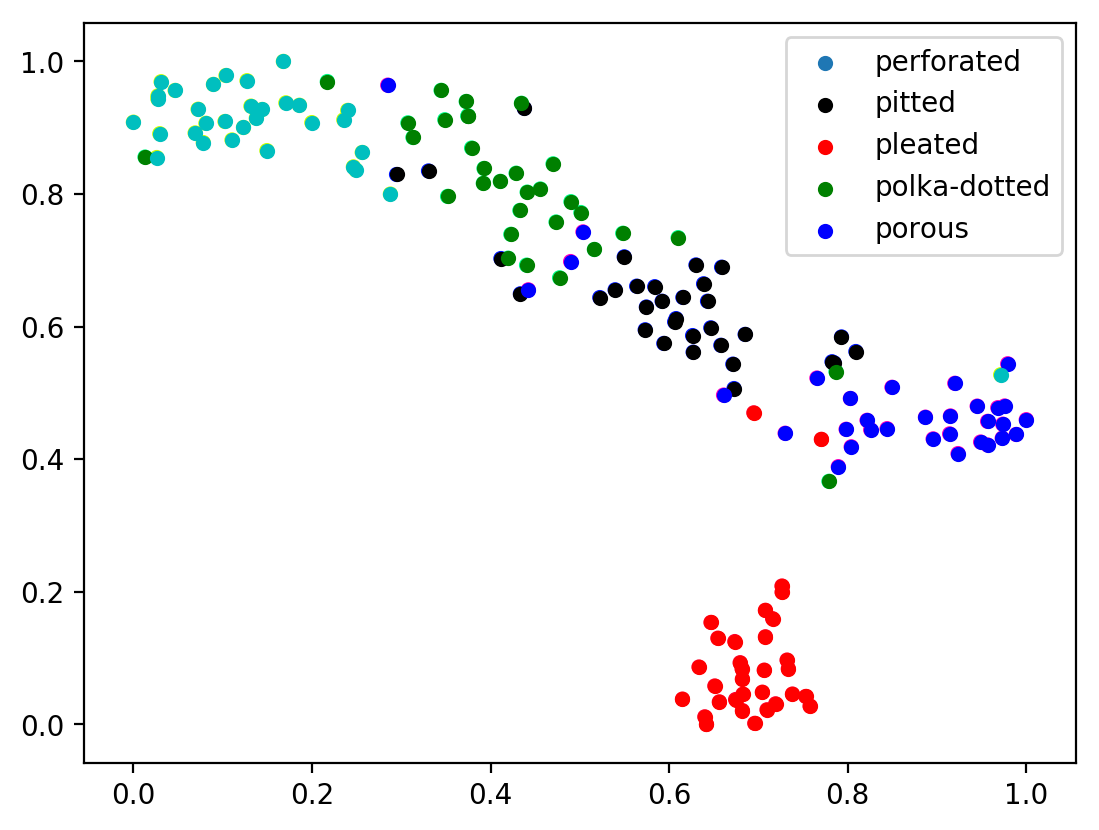}
			%\caption{fig2}
		\end{minipage}%
	}%
	\centering
	\caption{Illustration of the role of global context. The abscissa represents the channel and the ordinate represents the respective weight between 0 and 1 in the two figures on the left. The figure on the right is visualized by t-SNE dimensionality reduction. See detailed explanation in section \ref{subsection2}.}
	\label{fig:env2} 
\end{figure}

\begin{figure*}[htbp] 
	\centering  
	\includegraphics[width=150mm]{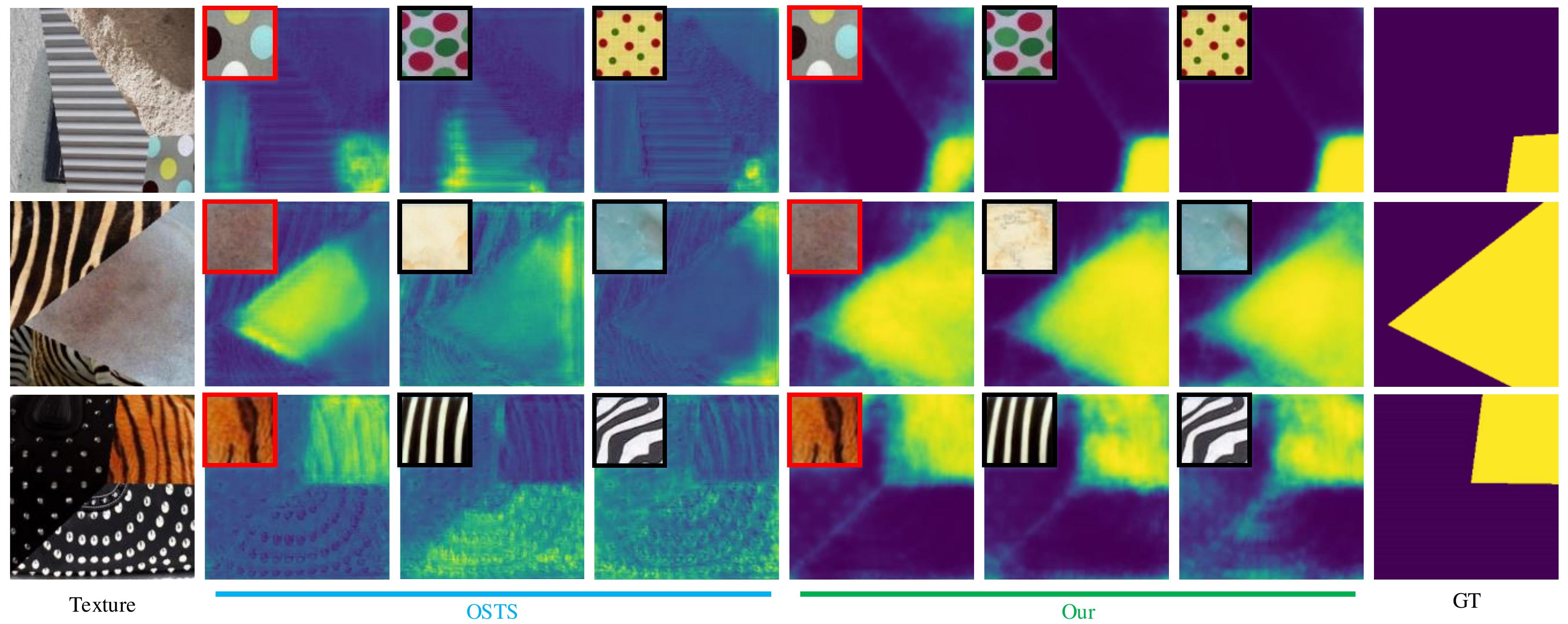}\\  
	\caption{Comparison of our segmentation results with OSTC. Three rows represent three groups of results. From left to right, they represent query image, results of OSTC compared to ours and groud truth respectively. Reference textures are show in the upper-left corner of each image of results. The reference images with red borders are from the original query image, while reference images with black borders are from the same class textures in DTD dataset. }   
	\label{fig:env3}  
\end{figure*}

\begin{table}[t]
	\centering
	\setlength{\tabcolsep}{4mm}{
	\begin{tabular}{lccccc}
		\toprule[1pt]
		\toprule[1pt]
		Method  & $i=0$ & $i=1$ & $i=2$ & mean \\
		\midrule
		OSTC  & $28.0$  & $34.0$ & $33.6$ & $31.9$   \\
		OSLSM   & $57.5$  & $47.2$ & $42.6$ & $49.1$   \\
		Ours   & $\textbf{60.7}$ & $\textbf{52.8}$ & $\textbf{44.8}$ & $\textbf{52.8}$   \\
		\bottomrule[1pt]
		\bottomrule[1pt]
	\end{tabular}}
	\caption{Mean IoU(\%) of our model and other state-of-the-art methods.}
	\label{tab:plain3}
\end{table}

\begin{figure*}[t]
	\centering  
	\includegraphics[width=150mm]{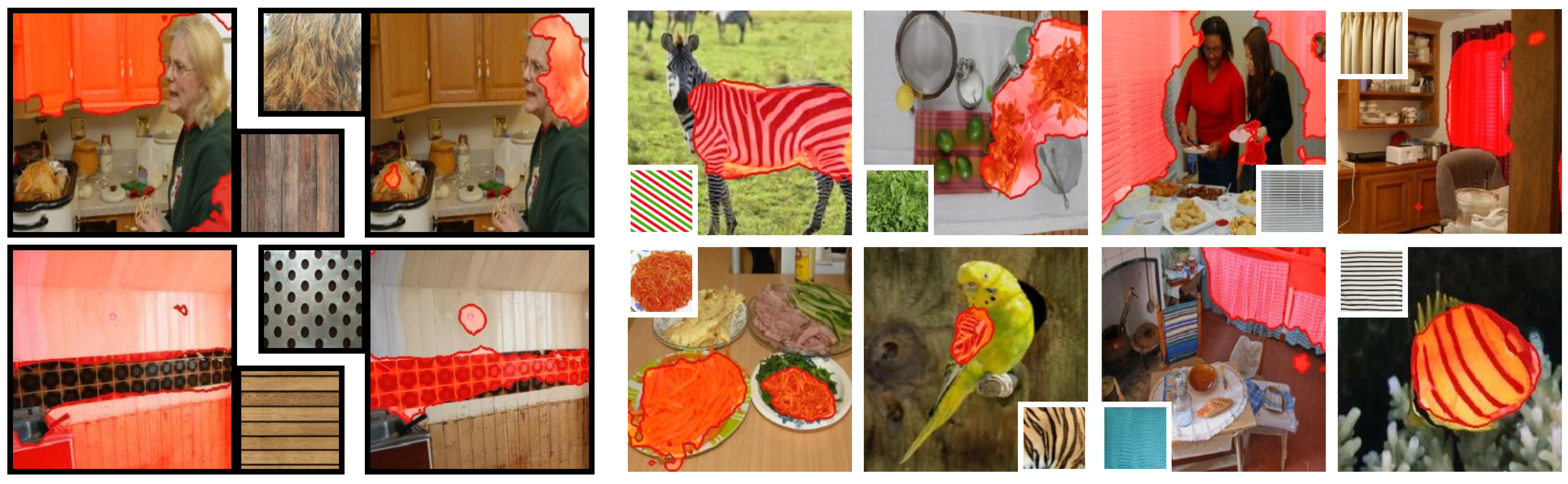}\\  
	\caption{Some qualitative results of our model in practical application. The reference images of each picture are placed in small graphics, and the corresponding segmentation results are marked in red.}   
	\label{fig:env6}  
\end{figure*}

\textbf{Effectiveness of Global Context Metric:} 
To demonstrate the function of the module, we visualize per-channel modulation weights for different reference images. We cite five categories in our test set as examples. As shown in two figures on the left in Figure \ref{fig:env2}, all the polylines of the same class in the left picture are almost identical and the right of five classes is different. To further explain this, we reduce the weights of these five classes and visualize them through t-SNE \cite{maaten2008visualizing}. It can be seen these five classes are clearly distinguished in the relation feature space. We think our module has learned an adaptive weight adjustment method among channels, where the intra-class is similar and inter-class appears inconsistent. As analyzed in Section \ref{subsection3}, this module combines global information to further separate different matching pairs in feature space.

\textbf{Comparision with state-of-the-art:} 
To better assess the overall performance of our reference network, we compare it to OSLSM and OSTC models. Because the tasks and model backbones are different, we set all the backbones to ResNet for fair comparision, and reproduce them with pytorch according to the two articles. All models are trained and tested in the same steps to achieve the purpose of adapting to our task. OSLSM is the first solution to one-shot semantic segmentation task, which is similar to our task. So after modifying backbone, we train their models directly in our tasks and don't need to modify any dataset settings. To compare with OSTC, we change the reference image to a $64\times 64$ patch for training in our task.

As can be seen from Table \ref{tab:plain3}, our model has more than three points mean IoU boost over the best performing OSLSM. Since OSTS is essentially designed for the interactive texture segmentation task, it does not work well for one-shot segmentation task. So we reproduce the model they use to solve their interactive texture segmentation task, which follows the settings of their papers. The reference texture of our model is the same class as a texture area in query image, and of course it may be the same texture image. Conversely, OSTS is not necessarily able to segment textures of the same class. As shown in Figure \ref{fig:env3}, the reference texture image in our model can achieve better segmentation effect whether it is the original image of a texture region of the query image or a different image of the same class. It benefits from the texture characteristics and the distinction of different texture features acquired by our texture
module and global context metric respectively.

\textbf{Evaluation on real scenarios:} 
Our model not only performs well in synthetic texture images, but also has scalability in practical applications. To prove this, we replace the query iamges with ones taken from real scenarios. We download high-quality indoor scene pictures with rich texture information from OpenSurfaces dataset \cite{bell2013opensurfaces}, and select animal and plant pictures with describable texture features from the Internet. For reference images, we randomly select images with similar texture information from DTD dataset and Internet according to query image content. For example, a striped image from DTD is randomly selected as a reference for zebras. In testing the above natural images, we did not pre-train or adjust the model. The parameters are still fixed in the state trained in synthetic texture images. As shown in Figure \ref{fig:env6}, We still get good segmentation results. It can be seen that our model can really learn new texture properties quickly, which is helpful for comprehensive scene understanding.

\section{Conclusion}

In this paper, a novel one-shot texture segmentation network OS-TR is proposed. By using a directionality-aware module to perceive texture variations at each direction and adjusting local features with global context information, OS-TR extracts the pixel information related to given texture in query images. In addition, our model can be well generalized from synthetic images to real scenarios without any adjustment. Experimental results show that our model is superior in both performance and adaptability with respect to existing methods.

\appendix

\bibliographystyle{named}
\bibliography{my}

\end{document}